# Optimization of Project Scheduling Activities in Dynamic CPM and PERT Networks Using Genetic Algorithms

**M. Hanefi CALP**[*1], **M. Ali AKCAYOL**[2]

[1]Karadeniz Technical University, OF Technology Faculty, Software Engineering, Trabzon, Turkey
[2]Gazi University, Engineering Faculty, Computer Engineering, Ankara, Turkey

**Abstract:** Projects consist of interconnected dimensions such as objective, time, resource and environment. Use of these dimensions in a controlled way and their effective scheduling brings the project success. Project scheduling process includes defining project activities, and estimation of time and resources to be used for the activities. At this point, the project resource-scheduling problems have begun to attract more attention after Program Evaluation and Review Technique (PERT) and Critical Path Method (CPM) are developed one after the other. However, complexity and difficulty of CPM and PERT processes led to the use of these techniques through artificial intelligence methods such as Genetic Algorithm (GA). In this study, an algorithm was proposed and developed, which determines critical path, critical activities and project completion duration by using GA, instead of CPM and PERT techniques used for network analysis within the scope of project management. The purpose of using GA was that these algorithms are an effective method for solution of complex optimization problems. Therefore, correct decisions can be made for implemented project activities by using obtained results. Thus, optimum results were obtained in a shorter time than the CPM and PERT techniques by using the model based on the dynamic algorithm. It is expected that this study will contribute to the performance field (time, speed, low error etc.) of other studies.

# Dinamik CPM ve PERT Ağlarında Genetik Algoritmalar Kullanarak Proje Çizelgeleme Aktivitelerinin Optimizasyonu

**Özet:** Projeler; amaç, zaman, kaynak ve çevre olmak üzere birbirleriyle ilişkili birtakım boyutlardan oluşmaktadır. Bu boyutların kontrollü kullanılması ve etkin planlanması proje başarısını getirmektedir. Proje planlama süreci, proje faaliyetlerinin tanımlanması ve projedeki faaliyetler için zaman ve kaynak tahmini yapılması süreçlerini kapsar. Bu noktada, proje kaynak planlama problemleri, Program Değerlendirme ve Gözden Geçirme Tekniği (PERT) ve Kritik Yol Metodu (CPM) birbiri ardına geliştirildikten sonra daha fazla dikkat çekmiştir. Bununla birlikte, CPM ve PERT işlemlerinin karmaşıklığı ve zorluğu, bu teknikleri Genetik Algoritma (GA) gibi yapay zeka yöntemleri ile kullanmaya itmiştir. Bu çalışmada, proje yönetimi kapsamında şebeke analizi için kullanılan CPM ve PERT tekniklerinin yerine, GA kullanılarak kritik yol, kritik faaliyet ve proje tamamlanma süresini belirleyen bir algoritma önerilmiş ve geliştirilmiştir. GA kullanılmasının amacı, bu algoritmaların karmaşık optimizasyon problemlerin çözümünde etkili bir yöntem olmasıdır. Böylece, elde edilen sonuçlar kullanılarak gerçekleştirilecek proje faaliyetleriyle ilgili doğru kararlar alınabilmektedir. Nitekim, geliştirilen dinamik algoritma ile CPM ve PERT tekniklerinden daha kısa sürede optimum sonuçlara ulaşılmıştır. Çalışmanın diğer çalışmaların performans alanına (zaman, hız, düşük hata vb.) katkı sağlaması beklenmektedir.

*Corresponding author: mhcalp@ktu.edu.tr

## 1. Introduction

It is necessary to manage the sources properly in today where time is important and sources are limited. In the opposite case, undesired consequences may be encountered. Evaluation and use of these resources optimally can be possible by planning process from the beginning to the end of the project. This situation reveals the concept of project management. Project management is described in literature as follows: "planning and controlling of process in order to make project arrive desired targets in a best ways and in an effective manner" [1], "planning, programming (scheduling) and controlling of project activities to achieve project targets" [2], "planning, organizing, executing and controlling of common activities of financial and human resources used in order to achieve a certain desired result" [3]. The main aim of project management is to complete a project in the planned period with the minimum cost and at the wanted quality level [4].

As it is seen from the above-mentioned definitions of the project management, planning and controlling of projects to be carried out are significant effecting elements for success of the project. Network analysis is one of the most widespread methods in the planning and controlling of the project. Sources of complex projects can be managed more effectively by analyzing with networks that are modal illustrations of a series of events and activities. Network means schedule that occurs from needed activities and events to achieve the aim of the program, and shows connections and relations between activities and events for planning requirement. In this study, GA was applied for the project analyzing on the networks instead of PERT or CPM methods. PERT and CPM have been developed to help planning, shape and controlling of large scaled complex projects. The networks are usually large and detailed so this situation forces to use a computer program to find project completion period by applying these models [5, 6].

For this reason, today's complex and hard conditions lead to find new methods for rapid and easy solutions. It is possible to group these methods as analytical or heuristic models that used to solve the problems in the complex and hard conditions. However, successful results couldn't be obtained using analytical models since analytical models are not effective to solve difficult or complex problems. Heuristic or evolutionary models are more effective in this regard, as they are problem-dependent. Especially, use of easy computing and evolutionary algorithms instead of difficult optimization techniques have come into prominence [7]. Evolutionary algorithms are very effective to resolve search and optimization problems. It seem that utilizing evolutionary algorithms is a highly effective way of finding various effective solutions in the realized studies [8].

At this point, GAs from evolutionary approaches have begun to take an important place. GAs overcome from above disadvantages since they are genetically powerful. GAs are a search and optimization method based on natural selection principals. There are successful applications of GAs in fields such as function optimization, scheduling, mechanical learning, design, and cellular production. Generally, there are four main steps in optimization procedure of the GAs: selection, crossing, mutation and evaluation. GA optimization procedure performs an iterative selection strategy, applies crossing and mutation, and then evaluates suitability of chromosomes. In this procedure, population includes good solutions and at the end, it supplies the best optimal solution [9-12].

In this section, CPM and PERT methods that are the subject of study are briefly explained before going into details of the study.

### 1.1. Critical path method

In CPM method, fixed and defined processing times are accepted. This method also seems necessary identification of the missing elements of the network, and project completion duration, critical activities and critical path of the network when differences are ignored [5, 13]. Critical path is a path that has the longest total activity period that supplies project to be completed in the shortest time between starting and ending point on the project network. The importance of critical path is that this length determines the completion duration of the project, and a delay in the activities that constitute the critical path retard the project in the same amount. For this reason, this path is critical path, and activities that make up this path are also called critical activities.

### 1.2. Program evaluation and review technique

PERT method is established on probability estimation of operation times and completion duration of the project [5, 14]. The most important difference between CPM and PERT is estimation of operation time. PERT was designed for projects in which durations are unclear such as research and development projects. CPM and PERT are calculated with the same method. The only difference is that PERT makes three predictions for each activity as seen in the Fig. 1. Here, a represents optimistic time (if all goes well), b represents pessimistic time (if all goes wrong), and m represents approximate time (under normal conditions).

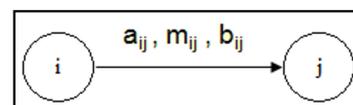

**Figure 1.** The displaying of the duration estimates for PERT Networks

The probability of occurrence of each activity within the estimated time is determined as a result of mathematical and statistical calculations done using these time estimates. The expected duration of activity (mean time) is calculated by the formula in Eq. 1.

$$T = (a+4m+b)/6 \qquad (1)$$

After durations of activity reduces to single time by using formulas, $E_i$ and $L_i$ values of each node are calculated from the first node to the last node. The nodes that have same E and L values give the critical path, the activities that are performed in this path give critical activities, and total length of the path give also project completion duration [15].

The study consists of four sections. In the first section of the article, a general introduction, and brief information about CPM and PERT were given. In the second section, related studies were given. In the third part, design and application of the model developed (material, method, structure and process) using genetic algorithm and, finally, in the fourth section, conclusions and future study were explained in detail.

**2. Related Studies**

There are many studies on network analysis within the context of project management. In this section, well-known and current studies were given. For example, Leu et al. suggested a new optimal construction time/cost trade-of method based on genetic algorithm by using network models for time-cost analysis on construction projects. Results demonstrated optimum adjustment of time and cost under different risk levels [16].

Azaron et al. suggested a genetic algorithm for the time-cost analysis on PERT networks. The study was applied on the only PERT networks by taking into account Erlang distributed activity time and direct costs. Researchers performed a few factorial experiments to specify suitable genetic algorithm parameters that generate optimal results within a given performing duration in the three typical status with different configurations. In addition, they compared genetic algorithm results with discrete-time approximation method results [17].

Baradaran et al. developed a hybrid distributed search approach for source-limited project planning problems on PERT type networks. Solutions was compared with optimal solution for small networks to validate the performance of new hybrid metaheuristic algorithm. Results demonstrated that suggested algorithm is suitable for small networks and other real problems [18].

Yakhchali suggested path-sorting algorithm for analysis of critical activities on CPM/PERT networks. In this study, it was intended to find the most suitable critical activities and paths by using fuzzy activity time values. In addition, an effective algorithm that based on a path enumeration technique was suggested to calculate the probability criticality degrees of activities. Performance of the algorithm was evaluated by using real project networks [19].

Pan et al. realized a study on planning project optimization by using Tabu Search Algorithm. In this study, results of developed algorithm and other artificial intelligence techniques were compared, and optimum parameters' settings were proposed for the models that are suggested through sensitivity analysis. In addition, suggested method provided a good user interface [20].

Abbasi and Mukattash analyzed solution of PERT network by using mathematical programming. In the study, it was aimed to minimize pessimistic time estimation of activities on the critical path. Suggested model showed that minimizing pessimistic time reduces project time. In addition, results of the model indicated that possibility of realizing the end node increased [21].

Haixiang et al. studied on a smart optimization to use for planning a coal-mining project in China. The optimization techniques they used include genetic algorithm, part bulk optimization and regular part bulk optimization. They handled the study in two stages. In the first stage, PERT network was created. In the second stage, they worked up to conclude by assigning estimated operation time of the activities to arrows [22].

Hashemin et al. made a study that optimizes allocation of limited nonrenewable source on PERT networks with different activity times. In the study, dynamic programming was utilized, and the aim is to maximize project completion duration without skipping any activity. The duration of each activity were categorized as either reducible or irreducible depending on network structure. Authors presented an analytical algorithm for each network structure. The algorithms were demonstrated by giving some examples [23].

Ke and Liu studied on solving project planning problem with mixture of randomness and fuzzy. In the study, it was aimed to adjust total cost and project completion duration by using genetic algorithm and fuzzy logic with available sources. Fuzzy simulations were given for unclear functions and they were embedded into genetic algorithm to develop a hybrid intelligent algorithm. As a result, some numerical experiment results were given for demonstration of performance of the algorithm [24].

Baar et al. suggested Tabu search algorithm for limited source project planning problems. The aim of the study is to define a schedule with minimal completion duration. Finally, a column-generation approach for a linear programming-based lower bound was presented and computational results were reported [25].

Demirel et al. solved problem of multistage integrated logistic network optimization with hybrid genetic algorithm. In their study, a mixed integer linear programming model that has limited capacity, multi-staged, and multi-product was developed for a general integrated logistic network design [26].

Kumar, firstly said that source management supplies that a project should be completed on time, at cost; and later source shortage is an expected cause for project delays. Suggested method, activity selected first for changing is based on the largest value of source rate. This procedure is repeated for all current activities for probable shifting of resources by searching optimal solution with GA. Developed GA model for project planning of sources has resulted in optimized output with reduced cost [27].

Rajeevan and Nagavinothini said that resource-constraint project scheduling problem is one of the very hard problems in Operations Research and it has lately become a popular area for the newest optimization models or techniques. For the resource-constraint project scheduling, meta-heuristic algorithm of optimization supplements conventional CPM technique and MSP (Microsoft Office Project) software for scheduling by optimizing project time. Results of this study showed that the algorithm is capable to meet best known solutions from problem space [28].

Hussain stated that resource allocation and leveling are two top challenges in construction project management, due to complex nature of construction projects, and CPM and PERT is not capable of minimizing undesirable fluctuations in resource utilization profile. In the study, researcher obtained results an optimum set of tasks and priorities that generate better-leveled resources profiles using Genetic Algorithm technique in MATLAB software [29].

Chitra and Halder aimed to develop a model that finds a proper trade-off between duration and cost to accelerate execution process in this study. He said that CPM is utilized to determine the longest time and cost needed for finalizing the project, and then the time-cost trade–off problem is formulated as a linear programming model. Karmake used LINDO program to define solution of the model. Data of proposed model were collected through interviews and direct discussion with the project managers of Chowdhury Construction Company, Dhaka, Bangladesh. Analysis results demonstrated that reduction of project duration by 17% is achieved by increasing cost by 3.73%, which is satisfactory [30].

Studies in the literature were summarized in Table 1 due to a better understanding.

Difference of this study from others, is that it can be applied on both CPM and PERT networks for the first time, and network models used in this study are dynamic. In addition, critical path of the project, critical activities and project completion duration can be obtained by entering desired number of nodes and activities through proposed model.

## 3. Design and Implementation of the Algorithm Developed by Using Genetic Algorithm

This study handles network analysis for project management on CPM and PERT networks. It searches solving of problems to be encountered in case network is very large during network analysis. The aim of this study is to solve network analysis problems by using GA, and to show that such complex problems can be solved by intuitional methods. In this study, "Genetic Algorithm" that is within heuristic methods were widely analyzed, and a new algorithm which is suitable for this strategy was developed and proposed.

### 3.1. Genetic algorithm

GA is the first artificial intelligence technique inspired by basic elements and phenomena of the Theory of Evolution brought to scientific world by Charles Robert Darwin. A typical GA is generally based on subjecting a variety of genetic manipulations according to fitness values of individuals represented by chromosome, which are to be initially used for solution and are represented as chromosomes in a selected coding. According to results of initial population, processes such as crossing and mutation are carried out. New individuals are obtained and created through relatively better individuals, and each new generation is examined until desired result is obtained or a certain stop criterion is obtained. In conclusion, GAs produce resolutions to optimization problems utilizing methods of natural evolution (inheritance, mutation, selection, and crossover). GA starts with creation of population. Solutions are obtained from population and utilized to form a new population. This process continues until the best solution is produced or until the number of population is determined [31-38].

In Table 2, general pseudo-code of GA was given [36, 38].

### 3.2. Material and method

Developed application to supply optimization with GA on CPM and PERT networks within the context of project management was briefly named as "GACPMPERT". Delphi 7 programming language was used for coding the designed algorithm. All simulations were practiced on the computers that have Intel Core I7 3.40 GHz processor. Algorithm was designed to be applied for all CPM and PERT networks. In the application, first of all nodes used on CPM and PERT networks, and then predecessor-successor status of mentioned nodes are determined. Operation durations of the activities are transferred to table according to above status. User inserts operation durations in order to use in CPM networks.

**Table 1.** Summary of the literature

| No | Author(s) Name | Problem | Method(s) | Result(s) |
|---|---|---|---|---|
| 1 | Leu et al. [16] | Time-cost analysis on construction projects. | A fuzzy optimal model based on genetic algorithm | Optimum adjustment of time and cost under different risk levels |
| 2 | Azaron et al. [17] | The time-cost analysis on PERT networks | Genetic algorithm and discrete-time approximation method | Optimal results within a given implementation duration in the three typical status with different configurations |
| 3 | Baradaran et al. [18] | Source-limited project planning problems on PERT type networks | A hybrid distributed search approach | Suitable results for small networks and other actual problems |
| 4 | Yakhchali [19] | Analysis of critical activities on CPM/PERT networks | Path-sorting algorithm | Most suitable critical activities and paths |
| 5 | Pan et al. [20] | Planning project optimization | Tabu search algorithm | Optimum parameters |
| 6 | Abbasi and Mukattash [21] | Solving PERT networks | Mathematical programming | Minimizing the pessimistic duration decreases project time and its variance and increase of the possibility of realizing the terminal node. |
| 7 | Haixiang et al. [22] | Planning a coal-mining project | A smart optimization | Optimal results |
| 8 | Hashemin et al. [23] | Allocation of limited nonrenewable source on PERT networks with different activity times | Dynamic programming | Optimum source allocation |
| 9 | Ke and Liu [24] | Solving project planning problem | Genetic algorithm and fuzzy logic | Successful arranging of total cost and project completion duration |
| 10 | Baar et al. [25] | Limited source project planning problems | Tabu search algorithm | Minimal completion duration |
| 11 | Demirel et al. [26] | The problem of multistage integrated logistic network optimization | Hybrid genetic algorithm | Minimum cost that satisfying the demands of customers |
| 12 | Kumar [27] | Possible shifting of resources by searching the optimal solution | Genetic algorithm | Optimized output with reduced cost |
| 13 | Rajeevan and Nagavinothini [28] | The resource constraints project scheduling problem | Meta-heuristic algorithm | Best known solutions from the problems space |
| 14 | Hussain [29] | The resource allocation | Genetic algorithm | Optimum set of tasks and priorities that generate better-leveled resources profiles |
| 15 | Chitra and Halder [30] | Finding a proper trade-off between time and cost to expedite the execution process | Developed model using LINDA | Reduction of project duration by 17%, increasing cost by 3.73% |

**Table 2**. Pseudo-code of the GA

> **Step 1.** Create population with N individuals according to target problem and selected coding scheme.
>
> **Step 2.** Iterative steps (for each individual and each purpose function size):
>
>    **2.1.** Step: Calculate the fitness function value with individual.
>
>    **2.2.** Step: Select individuals who will enter reproductive process.
>
>    **2.3.** Step: Apply crossover process to selected individuals.
>
>    **2.4.** Step: Apply mutation to some individuals.
>
> **Step 3.** Optimal solution at the end of iterative process is global best position and value(s).

In that case, in PERT networks, after optimistic, pessimistic, and approximate times are inserted, operation durations are calculated by using Equality 1 given in first part. In this point, application also presents a visual possibility by preparing a scenario CPM and PERT networks depending on different project situations entered. Later, genetic algorithm assumptions (size of population, elitism rate, generation number, and iteration number) for optimizing can be inserted dynamically depending on users' desire. However, only crossover is prepared to be applicable for all entered GA assumptions. As a result, genetic algorithm phases were completed rapidly and easily to optimize values and to reach optimal solution. Flow diagram of proposed algorithm is given at Fig. 2.

### 3.3. Structure and functioning of the algorithm

Developed and proposed algorithm can be applied on dynamic CPM and PERT networks. The algorithm was explained systematically in the frame of GA. Applying of suggested GA on any project was tried to be explained through a simple network example given in Fig. 3. Application was showed on CPM network diagram. The only difference of PERT networks is that operation duration is determined after calculating optimistic, pessimistic, and approximate duration.

*Chromosome and Gene*: It shows each solution of problems desired to be solved by genetic algorithm. In this study, genes and chromosomes were constituted depending on number of nodes entered by user. Each node can go their neighbor nodes one way and without skipping. Each node constitutes a gene, and paths formed by nodes constitute chromosomes. For example, in the Fig. 3, there are 11 nodes and 8 different paths. It means that there are 11 genes, and 8 chromosomes.

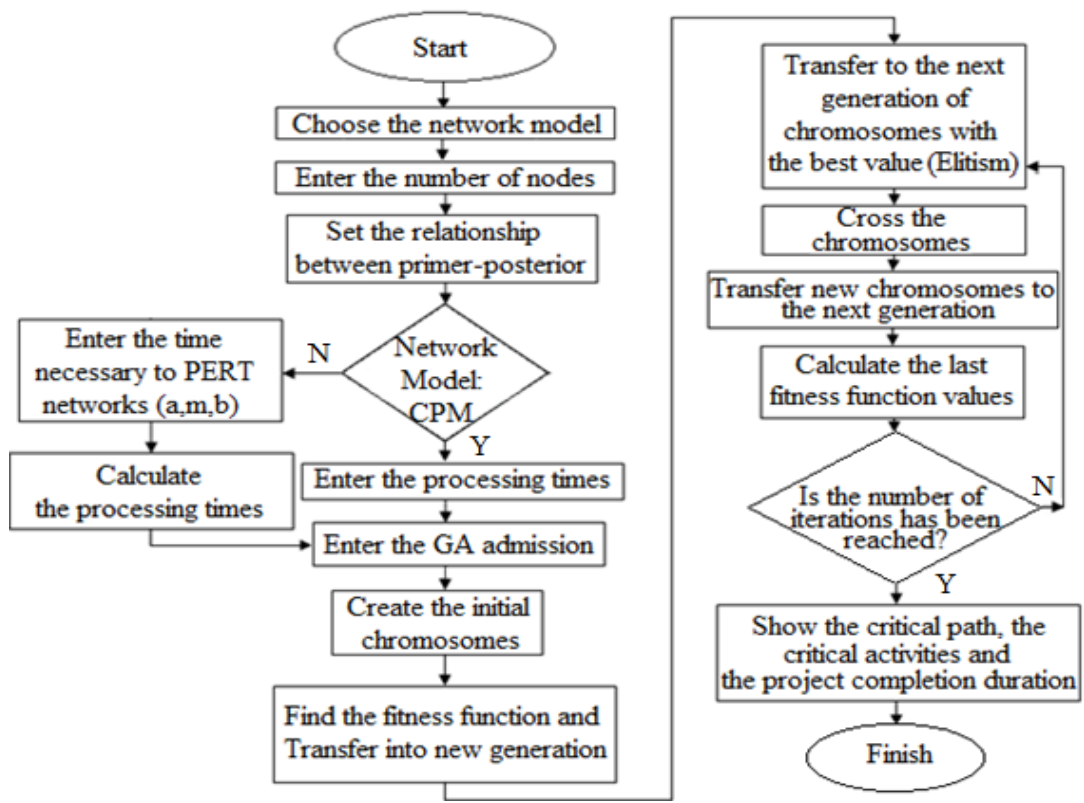

**Figure 2.** Flowchart of the proposed algorithm

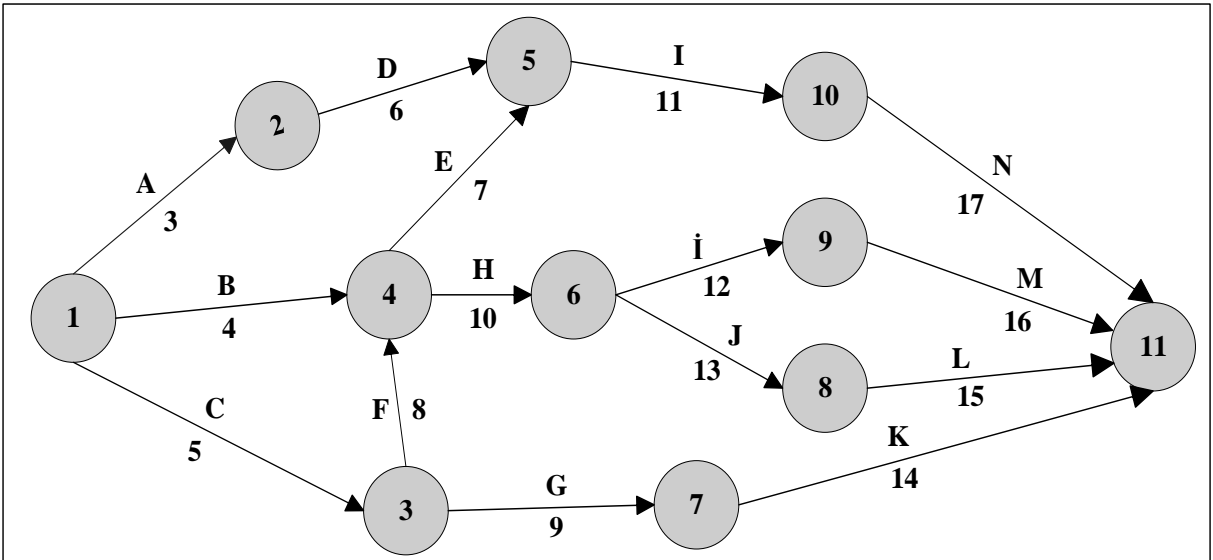

**Figure 3.** The example of network

```
for i:= 1 to satirsayisi do
    begin
        s:= sutunsayisi;
        r:=1;
        GotoLabel:
        rastgelesayi:= RandomRange(s+1,1);
        k:= Grid[rastgelesayi,r]; // Two-dimensional data
        if (k=0) then
            Goto GotoLabel // Conditionally exit the loop
        else
            Grid[r,i]:=k;
            r:=rastgelesayi;
         if r<>sutun then // If the 'r' value is not equal to the 'sutun' value
         Goto GotoLabel;
    end;
```

**Figure 4.** Algorithm that was developed for creating population

*Initial Population*: The set of solutions used to search for the best solution of the problem and randomly determined solutions were determined by the randomly generated set. The code line that was developed for initial population is given in the Fig. 4.

The entered nodes paths are taken into account in the process of population creation. The cells which is diagonal in the Fig. 5 should be zero (0) since CPM and PERT networks are one-way. The fact that the cells above the diagonal have zero (0) value indicates that there is no direct connection between the nodes. For example, the node 1 has a direct connection to the nodes 2, 3, and 4; however, it has not any direct connection to other nodes (nodes 5-11) since its cell value is zero. In the Fig. 6, a simple example was given for this situation.

In this way, direct connection status of each node is controlled one by one. Thus, the paths that form initial population were constituted. When the example in the Fig. 3 was taken into account, all paths (chromosomes) that form initial population were given in the Fig. 7, and some of the possible solutions were also given in the Fig. 8 with operation duration.

*Crossing*: "Single Point Crossing" technique was used in the developed application. In this application, a single cut-point was selected and it was obtained from the first ancestor chromosome till the cut-point. Offspring chromosome was created by combining the remaining part with the next part of the second ancestor's cut-point. Cut-point mentioned in crossing over was obtained from one third of length of the chromosome (path length). In case the length of the chromosome was not threefold, the obtained result was rounded up and then crossovered. A crossing example applied by algorithm developed on individuals created by the paths was given in the Fig. 9 and 10.

|  | D1 | D2 | D3 | D4 | D5 | D6 | D7 | D8 | D9 | D10 | D11 |
|---|---|---|---|---|---|---|---|---|---|---|---|
| D1 | 0 | 3 | 5 | 4 | 0 | 0 | 0 | 0 | 0 | 0 | 0 |
| D2 | 0 | 0 | 0 | 0 | 6 | 0 | 0 | 0 | 0 | 0 | 0 |
| D3 | 0 | 0 | 0 | 8 | 0 | 0 | 9 | 0 | 0 | 0 | 0 |
| D4 | 0 | 0 | 0 | 0 | 7 | 10 | 0 | 0 | 0 | 0 | 0 |
| D5 | 0 | 0 | 0 | 0 | 0 | 0 | 0 | 0 | 0 | 0 | 11 |
| D6 | 0 | 0 | 0 | 0 | 0 | 0 | 0 | 13 | 12 | 0 | 0 |
| D7 | 0 | 0 | 0 | 0 | 0 | 0 | 0 | 0 | 0 | 0 | 14 |
| D8 | 0 | 0 | 0 | 0 | 0 | 0 | 0 | 0 | 0 | 0 | 15 |
| D9 | 0 | 0 | 0 | 0 | 0 | 0 | 0 | 0 | 0 | 0 | 15 |
| D10 | 0 | 0 | 0 | 0 | 0 | 0 | 0 | 0 | 0 | 0 | 17 |
| D11 | 0 | 0 | 0 | 0 | 0 | 0 | 0 | 0 | 0 | 0 | 0 |

**Figure 5.** The displaying in table of processing times among nodes

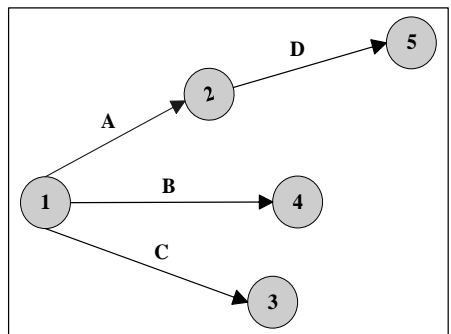

**Figure 6.** The displaying of the connections among nodes

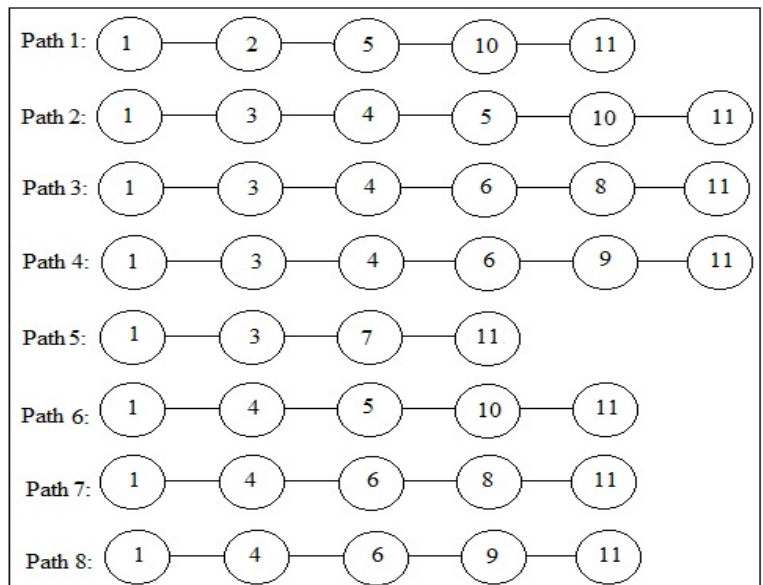

**Figure 7.** The chromosomes belong to initial population

|  | G1 | G2 | G3 | G4 | G5 | G6 | G7 | G8 | G9 | G10 | G11 |
|---|---|---|---|---|---|---|---|---|---|---|---|
| K1 | 5 | 0 | 8 | 10 | 0 | 13 | 0 | 15 | 0 | 0 | 0 |
| K2 | 5 | 0 | 8 | 10 | 0 | 12 | 0 | 0 | 15 | 0 | 0 |
| K3 | 4 | 0 | 0 | 10 | 0 | 13 | 0 | 15 | 0 | 0 | 0 |
| . | . | . | . | . | . | . | . | . | . | . | . |
| . | . | . | . | . | . | . | . | . | . | . | . |
| . | . | . | . | . | . | . | . | . | . | . | . |

**Figure 8.** Some of the possible solutions belong to chromosomes

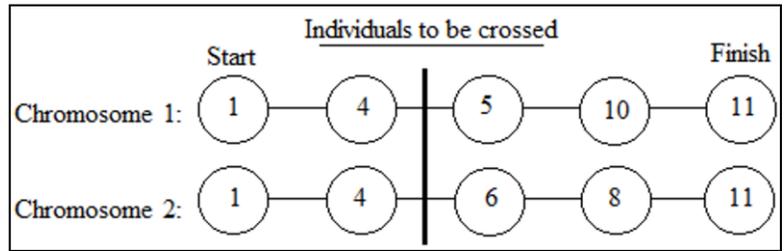

**Figure 9.** Crossover I

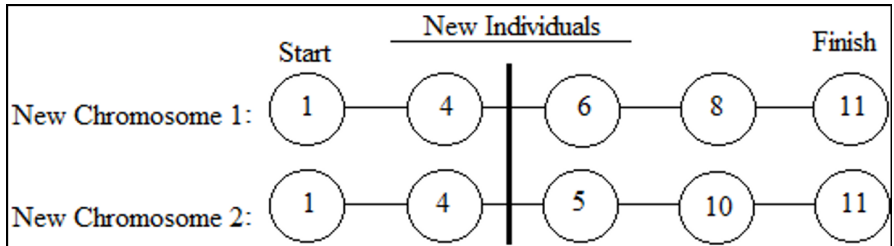

**Figure 10.** Crossover II

*Mutation:* It is applied in case some genetic information is lost in population and genes that create chromosome are all the same. Because same results were obtained in the study, there was no need to apply mutation on the application developed. It means that mutation did not affect the result of operation.

*Fitness Function (FF):* Fitness function used in the developed application was determined by detect the biggest operation duration in the paths after calculating sum of all one-way paths from the initial node of activities to the last node. The aim, here, was to obtain the path with the longest total activity time that ensure the project completed as soon as possible between the starting and ending point on the project network. Fitness function was given in Equality 2.

$$FF = \max (Y_{firstnode,j} + Y_{j,k} + Y_{k,l} + Y_{l,m} + Y_{m,n} + Y_{n,o} + \ldots + Y_{lastnode-1,lastnode}) \quad (2)$$

*Elitism:* Elitism method was used in this application despite the fact that different methods have been developed to create new population. In the study, the number of chromosome rate entered by user from generated population was selected as a best solution cluster, and others were omitted. Consequently, the best individuals were reached at each iteration, and thus the best optimum result was obtained at the last iteration.

*Displaying of the results:* All these genetic algorithm operations are repeated in equal amount to the number of iterations entered by the user and the gained optimum solution was presented to the user (Fig. 11).

In the study, for example, the values in Table 1 were entered for the activity and operation times of the network. GACPMPERT application reached the result by considering defined aim function depending on the available operation times.

Results were also demonstrated over the network with GACPMPERT application (Fig. 12). Path made by dashed lines indicates the critical path, activities on this road indicates critical activities, and sum of the durations on the critical path indicates project completion period.

Therefore, it can be seen clearly that the critical path is 1-3-4-6-8-11, critical activities (from the activities introduced between these paths) are C-F-H-J-L, and the project completion duration is 51 weeks (fifty-one).

The application was tested with five different projects data and all the results obtained were given in Table 4. According to Table 4, it was seem that accurate results were obtained in a short time using the proposed GACPMPERT application.

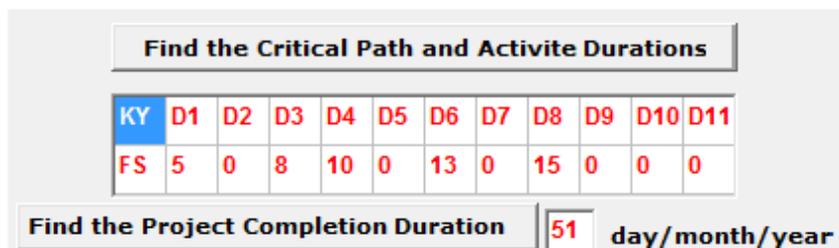

**Figure 11.** The displaying of the results

**Table 3.** The results of the GACPMPERT application example

| Line Number | Activite Name | Activite Duration |
|---|---|---|
|  | **START*** | **0** |
| 1 | A | 3 |
| 2 | B | 4 |
| **3** | **C*** | **5** |
| 4 | D | 6 |
| 5 | E | 7 |
| **6** | **F*** | **8** |
| 7 | G | 9 |
| **8** | **H*** | **10** |
| 9 | I | 11 |
| 10 | İ | 12 |
| **11** | **J*** | **13** |
| 12 | K | 14 |
| **13** | **L*** | **15** |
| 14 | M | 16 |
| 15 | N | 17 |
| **16** | **FINISH*** | **0** |
| Proje Duration | | 51 |
| Critical Path | | START (D1) - D3 - D4 - D6 - D8 - FINISH(D11) |
| Critical Activities | | C - F - H - J - L |

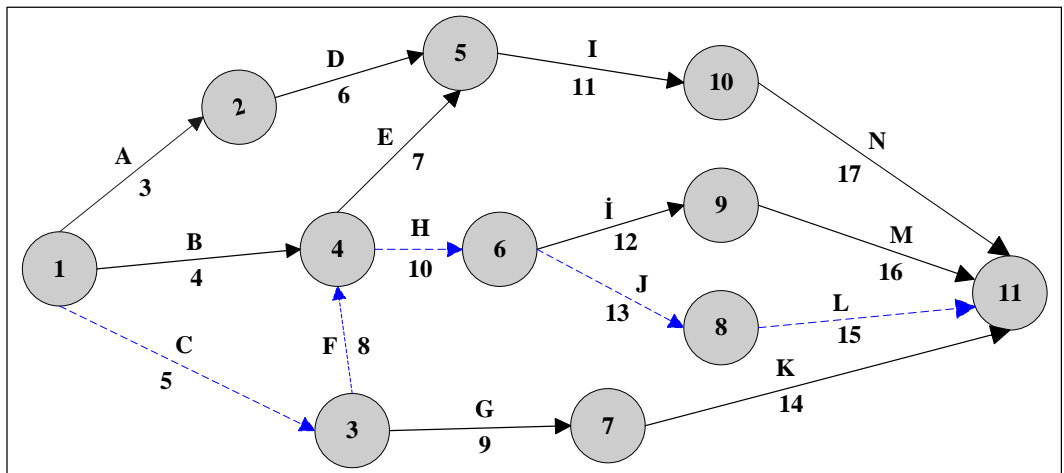

**Figure 12.** The displaying of the result on the network

## 4. Conclusions and Future Study

In this study, an algorithm was developed that finds the critical paths, critical activity, and project completion duration by using genetic algorithm instead of CPM and PERT techniques used for network analysis within the context of project management. Calculating critical path, critical activity, and project completion duration (without using GA) is a quite complex and long period in normal conditions. The algorithm is easy to implement for all CPM and PERT networks that are complex and time-consuming, too difficult to calculate manually. With the use of the developed software, any institution or organization will be able to determine how long it will complete the work to be performed by taking into account its own resources and constraints, which method to be used, and so they will be able to take measures depending on each new situation. Algorithm was developed for dynamic CPM and PERT networks, was tested using different GA acceptances, and the optimum results were obtained in a very short time.

In the next step of the study, all results obtained by applying hybrid approaches will be compared with the results obtained by applying GA algorithm developed in this study. Thus, more meaningful and productive results will be obtained and success of the proposed algorithm will reveal itself better.

**Table 4.** The results obtained depending on different inputs

| NO | | D1 START | D2 | D3 | D4 | D5 | D6 | D7 | D8 | D9 | D10 | D11 | D12 | D13 |
|---|---|---|---|---|---|---|---|---|---|---|---|---|---|---|
| ACTIVITY NAME | | | A | B | C | D | E | F | G | H | I | J | K | L |
| PUNITIVE ACTIVITIES | P1 (CPM) | - | - | - | A | A,B | B | C,D | E | - | - | - | - | - |
| | P2 (CPM) | - | - | A | B | G | D | A | C,F | D | A | D,I | - | F,K |
| | P3 (CPM) | - | - | A | B | B | H | D,E | A | C | E | A | I,J | - |
| | P4 (PERT) | - | - | A | B | B | C | B,C | D,E | C | - | - | - | - |
| | P5 (PERT) | - | - | - | A | A | D | B,E | B,C,E | F | G | H,I | - | - |
| DURATION (WEEK) | P1 (CPM) | 0 | 70 | 15 | 15 | 30 | 28 | 30 | 22 | - | - | - | - | - |
| | P2 (CPM) | 0 | 10 | 25 | 36 | 48 | 75 | 90 | 25 | 30 | 15 | 65 | 8 | 16 |
| | P3 (CPM) | 0 | 17 | 23 | 68 | 94 | 19 | 33 | 38 | 82 | 49 | 11 | - | - |
| | P4 (PERT) | 0 | 12 | 25 | 34 | 18 | 82 | 45 | 63 | 47 | - | - | - | - |
| | P5 (PERT) | 0 | 23 | 14 | 50 | 62 | 34 | 25 | 22 | 60 | 41 | 33 | - | - |
| PROJECT DURATION (WEEK) | P1 (CPM) | 130 | | | | | | | | | | | | |
| | P2 (CPM) | 190 | | | | | | | | | | | | |
| | P3 (CPM) | 258 | | | | | | | | | | | | |
| | P4 (PERT) | 191 | | | | | | | | | | | | |
| | P5 (PERT) | 215 | | | | | | | | | | | | |
| CRITICAL PATH | P1 (CPM) | START (D1) – D2 – D5 – D7 - FINISH (D9) | | | | | | | | | | | | |
| | P2 (CPM) | START (D1) – D2 – D3 – D4 – D9 – D6 – D7 – D13 - FINISH (D14) | | | | | | | | | | | | |
| | P3 (CPM) | START (D1) – D2 – D7 – D8 – D11 - FINISH (D12) | | | | | | | | | | | | |
| | P4 (PERT) | START (D1) – D2 – D4 – D6 – D8 - FINISH (D9) | | | | | | | | | | | | |
| | P5 (PERT) | START (D1) – D2 – D5 – D6 – DUMMY PATH– D8 – D10 – D11 – FINISH (D12) | | | | | | | | | | | | |
| CRITICAL ACTIVITIES | P1 (CPM) | A-D-F | | | | | | | | | | | | |
| | P2 (CPM) | A-F-G-D-J | | | | | | | | | | | | |
| | P3 (CPM) | A-B-C-H-E-F-L | | | | | | | | | | | | |
| | P4 (PERT) | A-C-E-G | | | | | | | | | | | | |
| | P5 (PERT) | A-D-E-DUMMY PATH-G-I-J | | | | | | | | | | | | |
| TIME (s) | P1 (CPM) | 5.4 | | | | | | | | | | | | |
| | P2 (CPM) | 5.7 | | | | | | | | | | | | |
| | P3 (CPM) | 7.3 | | | | | | | | | | | | |
| | P4 (PERT) | 7.8 | | | | | | | | | | | | |
| | P5 (PERT) | 8.2 | | | | | | | | | | | | |

**P:** Project
The time given in the PERT examples are the final values obtained from the calculation with the formula T = (a + 4m + b) / 6.